\documentclass[]{article}
\usepackage{proceed2e}
\usepackage{graphicx}
\usepackage{amsmath}
\usepackage{amssymb}
\usepackage{amsthm}
\usepackage[round]{natbib}
\usepackage{url}
\usepackage{multirow}

\DeclareMathOperator*{\argmin}{argmin}
\newtheorem{proposition}{Proposition}

\title{Generalized Fast Approximate Energy Minimization via Graph Cuts:\\$\alpha$-Expansion $\beta$-Shrink Moves} 

\author{ {\bf Mark Schmidt} \\  
INRIA - SIERRA Team\thanks{\ INRIA/ENS/CNRS UMR 8548}\\
Laboratoire d'Informatique\\
\'{E}cole Normale Sup\'{e}rieure\\  
Paris, France 
\And 
{\bf Karteek Alahari}  \\ 
INRIA - WILLOW Team\footnotemark[1]\\
Laboratoire d'Informatique\\
\'{Ecole} Normale Sup\'{e}rieure\\
Paris, France
} 
 
\begin{document} 
 
\maketitle 
 
\begin{abstract}
	We present $\alpha$-expansion $\beta$-shrink moves, a simple generalization of the widely-used $\alpha\beta$-swap and $\alpha$-expansion algorithms for approximate energy minimization.  We show that in a certain sense, these moves \emph{dominate} both $\alpha\beta$-swap and $\alpha$-expansion moves, but unlike previous generalizations the new moves require no additional assumptions and are still solvable in polynomial-time.  We show promising experimental results with the new moves, which we believe could be used in any context where $\alpha$-expansions are currently employed.
\end{abstract}
\section{Introduction}

We focus on the problem of  finding the most probable configuration in a pairwise Markov random field over discrete variables, 
a fundamental problem in the study of graphical models.  This is equivalent to minimizing the sum of a set of unary and pairwise energy functions defined over a set of discrete variables.    Problems of this type arise in many applications, but in general it is NP-hard to solve these problems even in the case of binary variables~\citep[see][Theorem 4.2]{kolmogorov2002energy}.  A classical method for computing an approximate solution to this problem is Besag's iterated conditional mode (ICM) algorithm~\citep{besag1986statistical}, but due to the local nature of this method it may become stuck in a poor local optimum.

In the special case where the variables are binary and the pairwise energy functions satisfy
a submodularity condition, it is possible to solve this problem in polynomial-time~\citep{hammer1965some,greig1989exact,kolmogorov2002energy}.  This has motivated the $\alpha\beta$-swap and $\alpha$-expansion moves proposed by~\citet{boykov1998markov,boykov1999fast}, which we review in \S\ref{sec:approximation}.  These methods find an approximate solution to a non-binary problem satisfying a submodularity condition over pairs of states (or triplets of states in the case of $\alpha$-expansions) by solving a sequence of binary submodular problems.  
The experiments of~\citet{szeliski2007comparative} show that
these algorithms perform well compared to competing approximate minimization methods at minimizing the energy functions arising from problems in computer vision, such as stereo image matching, creating photo montages, and image restoration.  Further, this study found that $\alpha$-expansions are often faster than competing methods with similar performance for these problems (such as variational message-passing algorithms like tree-reweighted belief propagation).  While originally introduced in the context of computer vision, these algorithms and their generalizations have also proved to be effective in other domains, such as protein structure prediction~\citep{gould2009alphabet}.

In a sense that we formally define in \S\ref{sec:dom}, $\alpha\beta$-swaps and $\alpha$-expansions \emph{dominate} the classic ICM algorithm.  However, although $\alpha$-expansions often lead to better experimental results than $\alpha\beta$-swaps~\citep{szeliski2007comparative}, under our definition of dominance neither of these more advanced methods dominates the other.  In this paper we propose a new type of move, $\alpha$-expansion $\beta$-shrink moves (\S\ref{sec:alphaBeta}).  These move are a simple generalization of both $\alpha\beta$-swap and $\alpha$-expansions, that \emph{dominates them both}.  Although we delay discussion of other generalizations of these moves to the end (\S\ref{sec:discussion}), we note that unlike previous generalizations the new moves require no additional assumptions beyond those needed to apply $\alpha$-expansions, and the moves can be computed in polynomial-time by solving a binary submodular problem defined on the original graph structure.  Thus, we can use them in place of $\alpha$-expansions for applications where these moves are currently used.  Our experiments on standard test problems from the field of computer vision (\S\ref{sec:experiments}) show that the new moves can lead to improved performance.

\section{Approximate Energy Minimization}
\label{sec:approximation}

The problem that we formally address is
\[
\max_{x \in \{1,2,\dots,N\}^p} p(x) \propto \prod_{i \in \mathcal{V}}\phi_i(x_i)\prod_{(i,j) \in \mathcal{A}}\phi_{ij}(x_i,x_j),
\]
where $\mathcal{V}$ and $\mathcal{A}$ are the vertices and arcs of a graph $(\mathcal{V},\mathcal{A})$, while the potentials $\phi_i(x_i)$ and $\phi_{ij}(x_i,x_j)$ map assignments of subsets of the discrete vector $x$ to non-negative values.
If we define the real-valued energy functions
\[
E_i(x_i) = -\log\phi_i(x_i), \quad E_{ij}=-\log\phi_{ij}(x_i,x_j),
\]
then finding the optimal assignment is equivalent to the following energy minimization problem:
\begin{equation}
\min_{x \in \{1,2,\dots,N\}^p} \sum_{i \in \mathcal{V}} E_i(x_i) + \sum_{(i,j) \in \mathcal{A}} E_{ij}(x_i,x_j),
\label{eq:pairwise}
\end{equation}
In general, solving this optimization problem is NP-hard~\citep[see][Theorem 4.2]{kolmogorov2002energy}, and a common approach to finding an approximate minimizer is with an iterative descent algorithm.  
The input to each iteration of these algorithms is a particular configuration of the variables $x$, and at each iteration an iterative descent method finds a configuration that minimizes the energy among a set $\mathcal{M}(x)$ of possible `moves', i.e. the next iteration is an element of
\begin{equation}
\label{eq:update}
\argmin_{y \in \mathcal{M}(x)}\sum_{i\in\mathcal{V}}E_i(y_i) + \sum_{(i,j)\in\mathcal{A}}E_{ij}(y_i,y_j).
\end{equation}
We have two conflicting desiderata on the set of moves $\mathcal{M}(x^k)$: we would like this set to be as large as possible, but we would like to able to efficiently find the optimal move.

\subsection{Iterated Conditional Mode}

In the classic ICM algorithm~\citep{besag1986statistical}, a node $j$ is selected and we replace $x_j$ with a value that maximizes the conditional probability $p(x_j|x_{-j})$, where $x_{-j}$ are the states of all variables except $j$.
In the framework of energy minimization, this can be viewed as an iterative descent method where the elements $y$ of the set of possible moves have the form
\[
y_i \leftarrow
\begin{cases}
\textrm{$\gamma \in \{1,2,\dots,N\}$} & \textrm{if $i=j$,}\\
\textrm{$x_i$} & \textrm{otherwise.}
\end{cases}
\]
That is, the current state of node $j$ can be replaced by any other possible state, a form of coordinate descent.  
We use $\mathcal{M}_j^I(x)$ to denote the set of all $y$ of this form.   With this definition of the move space, the iterative descent update~\eqref{eq:update} for the ICM move given $x$ and $j$ simplifies to
\begin{equation}
\label{eq:ICMenergy}
\argmin_{y \in \mathcal{M}_j^I(x)} E_j(y_j | x_{-j}),
\end{equation}
where we will find it convenient to define the \emph{conditional energy} of a variable $i$ given a set of variables $a$ as
\begin{equation}
\label{eq:condEnergy}
\begin{aligned}
E_i(x_i|x_a) & = E_i(x_i) \\
& + \sum_{j | j \in a, (i,j) \in \mathcal{A}} E_{ij}(y_i,x_j) \\
& + \sum_{j | j \in a, (j,i) \in \mathcal{A}} E_{ij}(x_j,y_i).
\end{aligned}
\end{equation}
While $\exp(-E_i(x_i|x_{-i})) \propto p(x_i|x_{-i})$, this is a slight abuse of the conditioning notation since for other conditioning sets $a$ it ignores factors that depend on variables besides $i$ and those in $a$.
Clearly, we can efficiently compute the optimal ICM move by simply testing each $E_j(y_j | x_{-j})$.

\subsection{$\alpha\beta$-Swaps}

We say that a pairwise energy function $E_{ij}$ defined on binary variables is \emph{submodular} if it satisfies the inequality\footnote{Submodularity is normally defined as a property of functions on sets, and its use here is because it is equivalent to submodularity of a function that takes the set of variables labeled $2$ and returns the corresponding $E_{ij}$\citep[\S7]{kolmogorov2002energy}.}
\begin{equation}
E_{ij}(1,1) + E_{ij}(2,2) \leq E_{ij}(2,1) + E_{ij}(1,2).
\label{eq:binarySubMod}
\end{equation}
This type of pairwise energy prefers the neighbouring variables to take the same state.  In the special case of binary variables where all pairwise energies are submodular, the optimal solution to~\eqref{eq:pairwise} can be computed in polynomial-time as a minimum-cut problem, see~\citep{kolmogorov2002energy}.\footnote{There also exist several other notable cases where it possible to compute the solution in polynomial-time, such as the case where the pairwise energies are convex~\citep{ishikawa2003exact}, or in the case of general energies where the graph structure has low treewidth~\citep[\S13]{koller2009probabilistic} or is outer-planar~\citep{schraudolph21efficient}.  However, these are not our focus and a full discussion of this extensive literature is outside the scope of the current work.}

In some non-binary problems we have, for all combinations of states $\alpha$ and $\beta$, that the pairwise energies $E_{ij}$ satisfy
\begin{equation}
E_{ij}(\alpha,\alpha) + E_{ij}(\beta,\beta) \leq E_{ij}(\beta,\alpha) + E_{ij}(\alpha,\beta).
\label{eq:multiSubMod}
\end{equation}
That is, when restricted to any two states $\alpha$ and $\beta$, the energy function is submodular.  
Though we can no longer guarantee that we can find the optimal solution in polynomial-time given only this restriction, it does allow us to take advantage of the ability to efficiently solve binary submodular problems in order to implement a more powerful descent move.

In particular, given two states $\alpha$ and $\beta$, the set of moves $\mathcal{M}_{\alpha\beta}^S(x)$ associated with the $\alpha\beta$-swap move introduced by~\citet{boykov1998markov} are of the form:
\[
y_i \leftarrow
\begin{cases}
\textrm{$\alpha$ or $\beta$} & \textrm{if $x_i = \alpha$ or $x_i = \beta$,}\\
\textrm{$x_i$} & \textrm{otherwise.}
\end{cases}
\]
That is, the move can simultaneously change any combination of nodes labeled $\alpha$ to $\beta$, and any combination of nodes labeled $\beta$ to $\alpha$.  
In this case, the iterative descent update is a solution of the problem
\begin{align*}
\argmin_{y \in \mathcal{M}_{\alpha\beta}^S(x)}  &\sum_{i \in \mathcal{V} | x_i \in \{\alpha,\beta\}} E_i(y_i|x_{-\alpha\beta}) \\
& +  \sum_{(i,j) \in \mathcal{A} | x_i,x_j \in \{\alpha,\beta\}}E_{ij}(y_i,y_j),
\end{align*}
where we again make use of our definition of the conditional energy~\eqref{eq:condEnergy} and where
we have used $x_{-\alpha\beta}$ to reference the states of the variables not labeled $\alpha$ or $\beta$.  
This is a binary problem over the subgraph induced by the nodes labeled $\alpha$ or $\beta$, and under condition~\eqref{eq:multiSubMod} this update can be computed in polynomial-time because all edges
in the induced subgraph are submodular.

\subsection{$\alpha$-Expansions}

A closely-related set of moves later proposed by~\citet{boykov1999fast} are $\alpha$-expansions.  Here, we choose a state $\alpha$ and we can use $\alpha$ to replace the current state of any variable.  Thus, this set of moves $\mathcal{M}_\alpha^E(x)$ is of the form
\[
y_i \leftarrow
\begin{cases}
\textrm{$\alpha$} & \textrm{if $x_i = \alpha$,}\\
\textrm{$\alpha$ or $x_i$} & \textrm{otherwise.}
\end{cases}
\]
We can write the optimal $\alpha$-expansion as the solution to the problem
\begin{align*}
\argmin_{y \in \mathcal{M}_{\alpha}^E(x)}  &\sum_{i \in \mathcal{V} | x_i \neq \alpha} E_i(y_i|x_{\alpha}) \\
& +  \sum_{(i,j) \in A | x_i \neq \alpha ,x_j \neq \alpha}E_{ij}(y_i,y_j),
\end{align*}
where we use $x_\alpha$ to reference variables labeled $\alpha$.
This is again a binary problem, this time on the subgraph induced by those nodes not labeled $\alpha$.  
However,
condition~\eqref{eq:multiSubMod} is no longer sufficient to guarantee that the edges in the induced subgraph are submodular since each pairwise term involves the three states $\alpha$, $x_i$, and $x_j$ (which may all be different).  Nevertheless, it is sufficient to ensure that 
\begin{equation}
E_{ij}(\alpha,\alpha) + E_{ij}(\gamma_1,\gamma_2) \leq E_{ij}(\gamma_1,\alpha) + E_{ij}(\alpha,\gamma_2),
\label{eq:triangle}
\end{equation}
for all combinations of states $\alpha$, $\gamma_1$, and $\gamma_2$.  This is a stronger condition than~\eqref{eq:multiSubMod}, which corresponds to the special case where $\gamma_1=\gamma_2$.  If $E_{ij}(\alpha,\alpha) = 0$ for all $\alpha$, this is the triangle inequality.

\begin{figure*}
\centering
\label{fig:alphaBeta}
\includegraphics[width=.5\columnwidth]{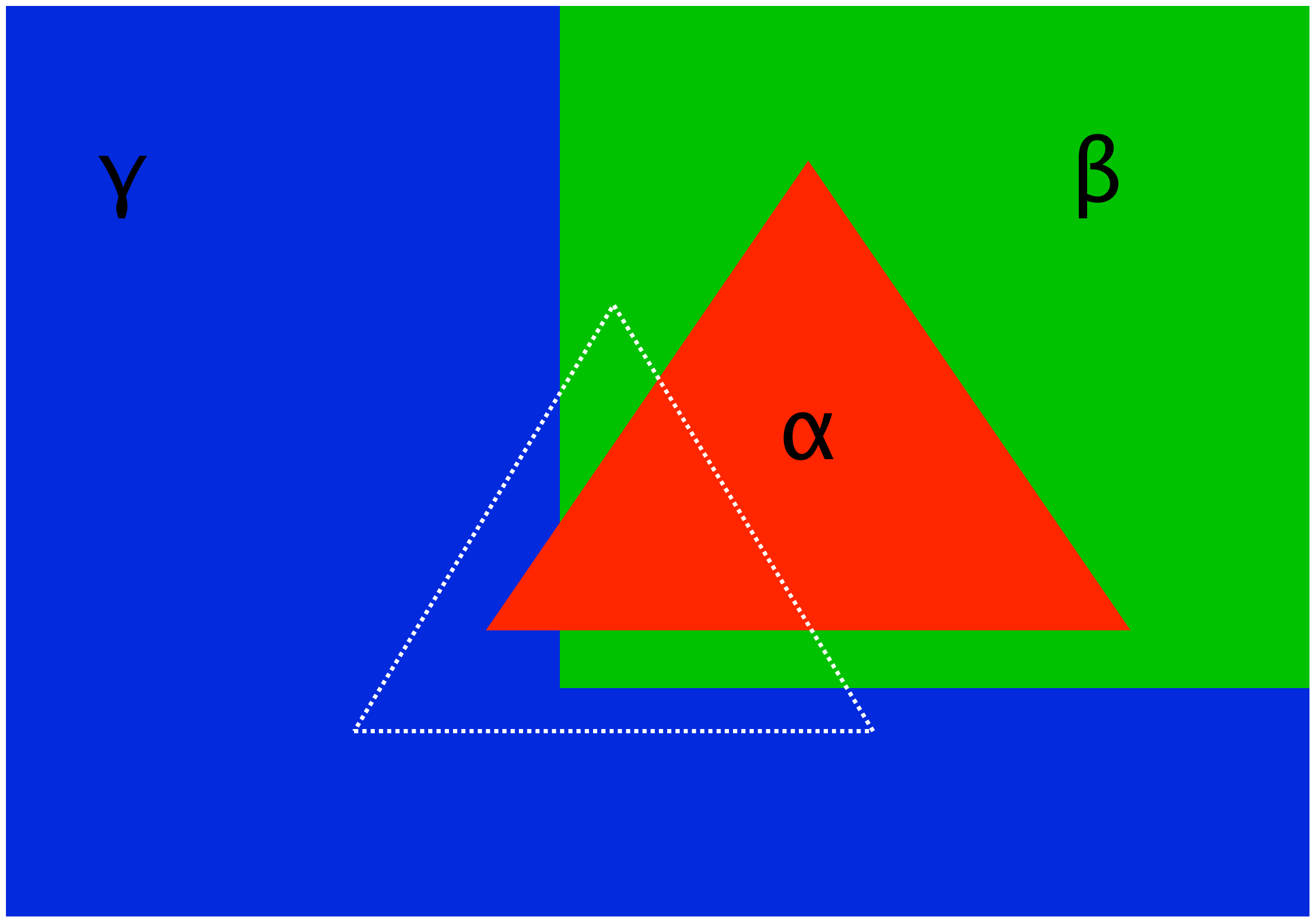}
\includegraphics[width=.5\columnwidth]{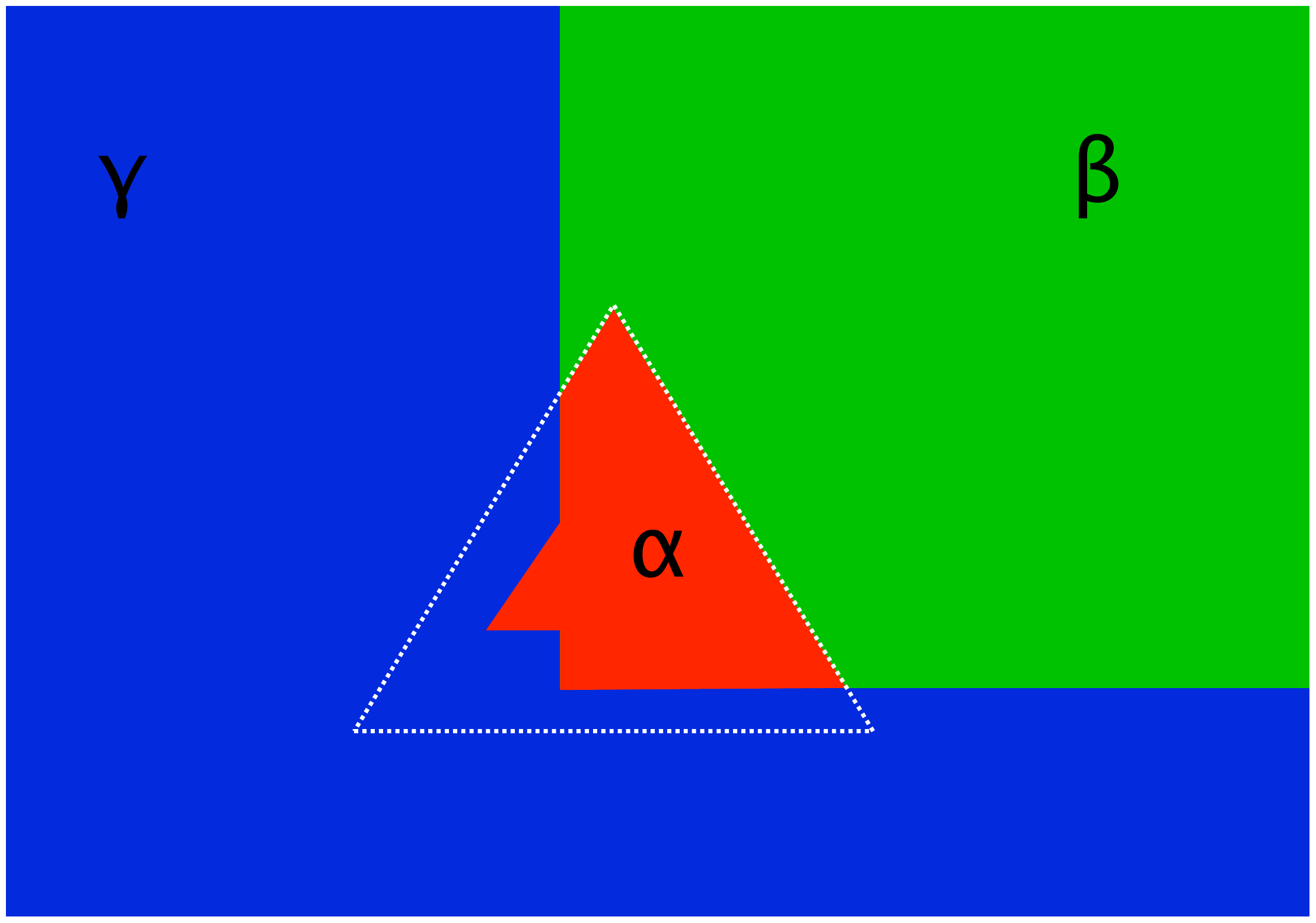}
\includegraphics[width=.5\columnwidth]{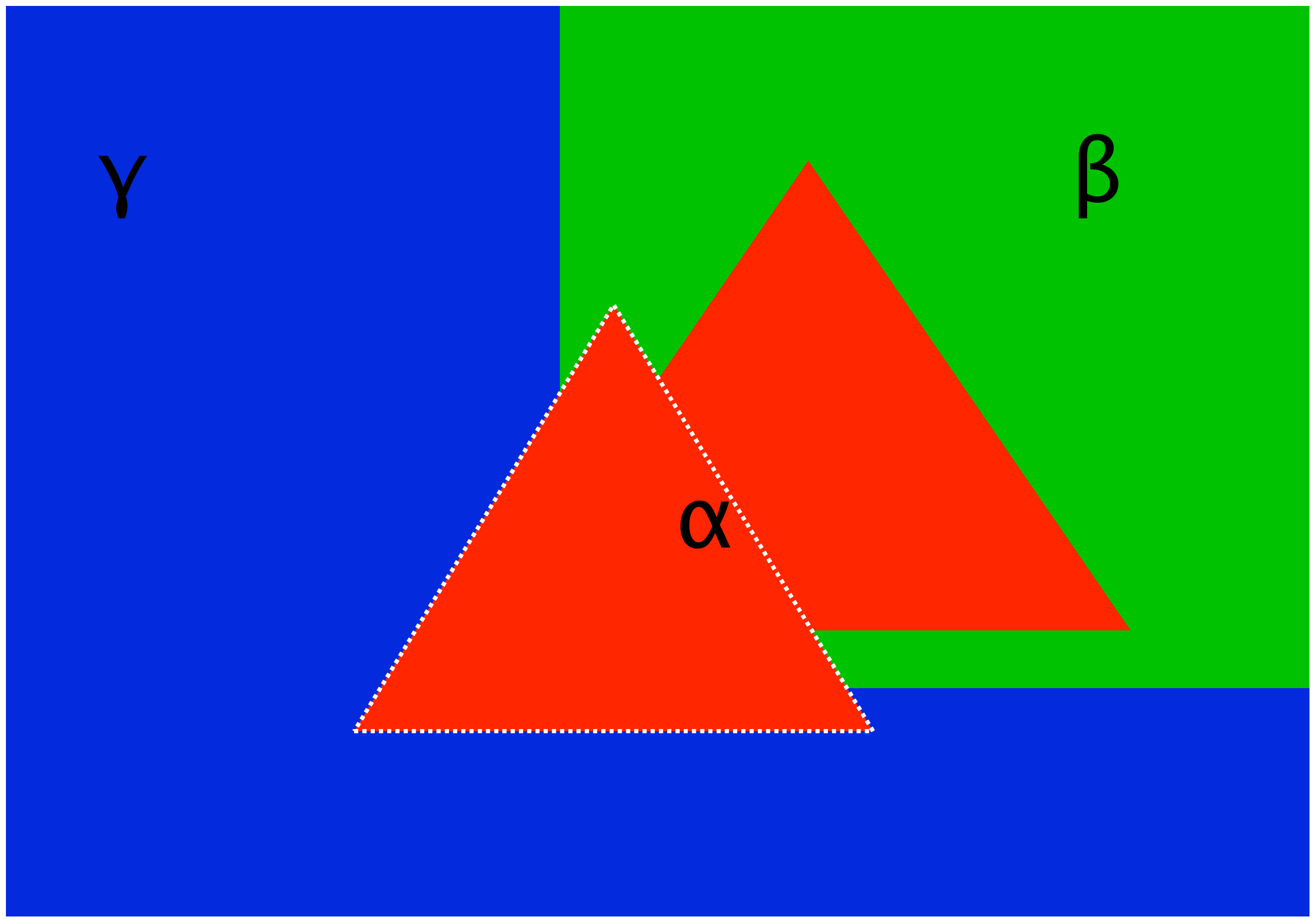}
\includegraphics[width=.5\columnwidth]{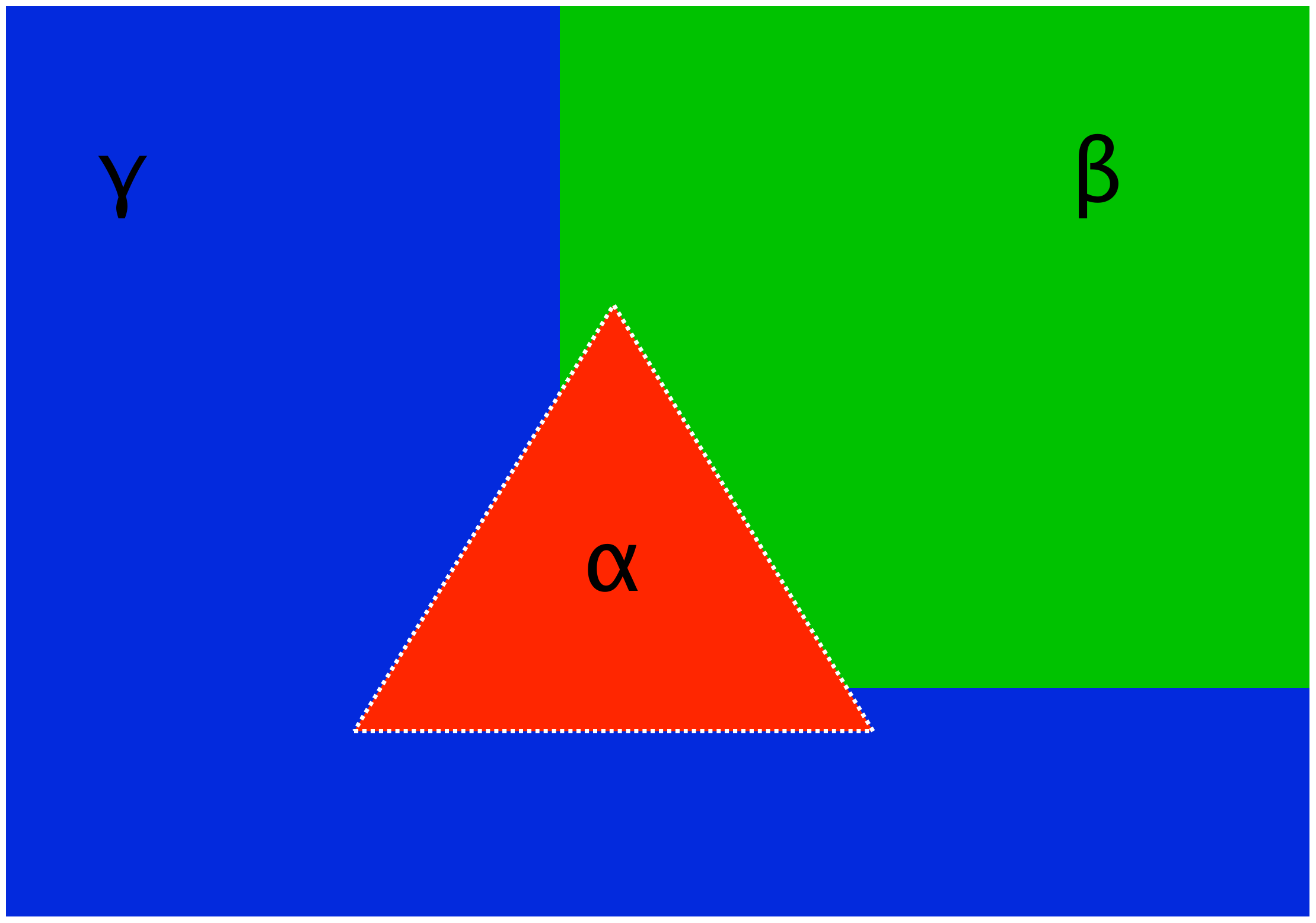}
\caption{From left to right: Initial labeling, labeling after $\alpha\beta$-swap, labeling after $\alpha$-expansion, labeling after $\alpha$-expansion $\beta$-shrink.  The optimal labeling of the $\alpha$ pixels is outlined by a white triangle, and is achieved from the initial labeling by one $\alpha$-expansion $\beta$-shrink move.}
\end{figure*}

\section{Local Dominance of Iterative Algorithms}
\label{sec:dom}

 
 Some authors analyze the approximation accuracy of local optima found by iterative descent methods such as $\alpha$-expansions~\citep[\S4.3.4]{veksler1999efficient}, but the guarantees on the approximation accuracy tend to be very conservative since these global bounds must hold over all possible optima.
Further, they often do not provide guidance in choosing between iterative descent methods.  For example, a small modification to the analysis of~\citep[see][\S4.3.4]{veksler1999efficient} gives us an identical global bound for the ICM algorithm.
 
 Instead of focusing on such a global analysis, we instead focus on a simple local analysis of the methods.  Specifically, we are interested in the improvement in the objective function that can be obtained in a single iteration with different types of moves, given the same input to the iteration.  From this viewpoint, we say that a move set $\mathcal{M}^A(x)$ \emph{dominates} a move set $\mathcal{M}^B(x)$ if:
\begin{enumerate}
\item For any energy function $E(x)$ and any input configuration $x$, optimizing over $\mathcal{M}^A(x)$ never gives a higher energy than optimizing over $\mathcal{M}^B$(x):
\begin{equation}
\label{eq:dom1}
\forall_{E,x}, \min_{y \in \mathcal{M}^A(x)}E(x) \leq \min_{y \in \mathcal{M}^B(x)}E(x).
\end{equation}
\item There exists energy functions $E(x)$ and input configurations $x$ where optimizing over $\mathcal{M}^A(x)$ gives a strictly lower energy than optimizing over $\mathcal{M}^B(x)$:
\begin{equation}
\label{eq:dom2}
\exists_{E,x}, \min_{y \in \mathcal{M}^A(x)}E(x) < \min_{y \in \mathcal{M}^B(x)}E(x).
\end{equation}
\end{enumerate}
This is an intuitive notion of the relative local improvement that is made by different moves, but it can also be used as a measure of the relative `strengths' of local optima found by different types of moves; if the move set $\mathcal{M}^A(x)$ dominates the move set $\mathcal{M}^B(x)$, 
then no move in the set $\mathcal{M}^B(x)$ will ever be able to improve on a local minimum with respect to move set $\mathcal{M}^A(x)$, but the optimal move in move set $\mathcal{M}^A(x)$ may be able to escape from a local minima with respect to move set $\mathcal{M}^B(x)$. 

To establish dominance relations between various search moves, it will be convenient to define versions of the moves that do not depend on any parameters except $x$.  Thus, we make the following definitions:
\begin{align*}
\mathcal{M}^I(x) & = \cup_j \mathcal{M}_j^I(x),\\
\mathcal{M}^S(x) & = \cup_{\alpha,\beta} \mathcal{M}_{\alpha\beta}^S(x),\\
\mathcal{M}^E(x) & = \cup_{\alpha} \mathcal{M}_{\alpha}^E(x).
\end{align*}
That is, these expanded move spaces simply search over all possible values of the parameter(s).  
Under these definitions, it is straightforward to establish the following relationships between the methods:
\begin{proposition}
The $\alpha\beta$-swap move set $\mathcal{M}^S(x)$ dominates the ICM move set $\mathcal{M}^I(x)$.  The $\alpha$-expansion move set $\mathcal{M}^E(x)$ dominates the ICM move set $\mathcal{M}^I(x)$.
\end{proposition}
\begin{proof}
Assume an arbitrary energy function $E$ and configuration $x$, and let $y$ be the optimal move in $\mathcal{M}^I(x)$, which must be an element of $\mathcal{M}_j^I(x)$ for some $j$.  Then the optimal ICM move is in $\mathcal{M}_{\alpha\beta}^S(x)$ with $\alpha = x_j$ and $\beta = y_j$, which establishes~\eqref{eq:dom1}.  
To establish~\eqref{eq:dom2}, it is sufficient to construct an energy function $E$ along with a configuration $x$ where the optimal solution involves changing two elements of $x$. For example, take a two variable problem with $x=(1,1)$ where the optimal solution is $(2,2)$.  Establishing that $\alpha$-expansions dominate ICM can be done similarly.
\end{proof}
\begin{proposition}
The $\alpha$-expansion move set $\mathcal{M}^E(x)$ does not dominate the $\alpha\beta$-swap move set $\mathcal{M}^S(x)$.  The $\alpha\beta$-swap move set $\mathcal{M}^S(x)$ does not dominate the $\alpha$-expansion move set $\mathcal{M}^E(x)$.
\end{proposition}
\begin{proof}
To establish that $\alpha$-expansions do not dominate $\alpha\beta$-swaps, it is sufficient to construct an energy function $E$ and configuration $x$ where the optimal solution is obtained by a single $\alpha\beta$-swap but can not be obtained by a single $\alpha$-expansion.  For example, take a two-variable problem with $x = (1,2)$ where the optimal solution is $(2,1)$.  We can similarly show that $\alpha\beta$-swaps do not dominate $\alpha$-expansions by taking a three variable problem with $x=(1,2,3)$ where the optimal solution is $(1,1,1)$.
\end{proof}

Thus, based on the definition of dominance discussed in this section, we should prefer both $\alpha\beta$-swaps and $\alpha$-expansions to ICM.  However, local dominance does not guide us in selecting between these two more advanced methods.

%

\section{$\alpha$-Expansion $\beta$-Shrink Moves}
\label{sec:alphaBeta}

Towards the goal of developing a method that dominates both these methods, consider a generalization of both $\alpha\beta$-swaps and $\alpha$-expansions where the set of moves $\mathcal{M}_{\alpha\beta}^G(x)$ are of the form:
\[
y_i \leftarrow
\begin{cases}
\textrm{$\alpha$ or $\beta$} & \textrm{if $x_i = \alpha$}\\
\textrm{$\alpha$ or $x_i$} & \textrm{otherwise}
\end{cases}
\]
That is, we allow any node not currently labeled $\alpha$ to take the value $\alpha$, but in addition we allow the nodes currently labeled $\alpha$ to take the value $\beta$ ($\alpha$ expands everywhere, but is shrunk by $\beta$).  We call these $\alpha$-expansion $\beta$-shrink moves, and Figure~\ref{fig:alphaBeta} illustrates an instance of the move.  

Under the definition of the previous section, we have the following result for an analogously defined $\mathcal{M}^G(x)$:
\begin{proposition}
The $\alpha$-expansion $\beta$-shrink move set $\mathcal{M}^G(x)$ dominates the $\alpha\beta$-swap move set $\mathcal{M}^S(x)$.  The $\alpha$-expansion $\beta$-shrink move set $\mathcal{M}^G(x)$ dominates the $\alpha$-expansion move set $\mathcal{M}^E(x)$.
\end{proposition}
\begin{proof}
To establish~\eqref{eq:dom1}, we note that $\mathcal{M}^G(x)$ includes all $\alpha\beta$-swaps and $\alpha$-expansions as special cases.  Further, it also includes moves that are not instances of either of these moves (i.e.\ cases where $\alpha$ expands into more than one state, but $\beta$ shrinks into $\alpha$).  Thus, to establish~\eqref{eq:dom2} we can simply take a three variable problem with $x=(1,2,3)$ where the optimal solution is $(2,1,1)$.  
\end{proof}
Note that the argument above can also be used to show that the new move set $\mathcal{M}^G(x)$ also dominates the move set $\mathcal{M}^S\cup\mathcal{M}^E$, the union of $\alpha\beta$-swap and $\alpha$-expansion moves.

We can write the optimal $\alpha$-expansion $\beta$-shrink move as a solution to the problem
\begin{equation}
\label{eq:genMove}
\argmin_{y \in \mathcal{M}_{\alpha\beta}^G(x)}  \sum_{i \in \mathcal{V}} E_i(y_i) +  \sum_{(i,j) \in \mathcal{A}}E_{ij}(y_i,y_j),
\end{equation}
Note that unlike the previous move sets we have discussed where the move is computed by solving a problem involving conditional energies over an induced subgraph, this move involves solving a problem with the original unary energies on the original graph structure.  In some scenarios, these properties might make the generalized move simpler to implement than $\alpha\beta$-swaps or $\alpha$-expansions.  Further, if the original graph has a special structure, it allows the use of specialized codes for solving binary minimum-cut problems with this structure.
For example,~\citet{delong2008scalable} propose a method for the special case of problems with very large grid structures.  The simple form of the subproblem may also simplify the implementation of \emph{dynamic} graph cuts \citep{kohli2007dynamic}, where the similarity between subproblems when using the same value of $\alpha$ (and $\beta$) is used to substantially speed up the computation.

There are many possible moves we could define that would dominate one or more of the  methods discussed in \S\ref{sec:approximation}, and we return to this topic in \S\ref{sec:discussion}.  However, an appealing property of $\alpha$-expansion $\beta$-shrink moves is the following:
\begin{proposition}
If for each edge $E_{ij}$ condition~\eqref{eq:triangle} holds for all $\gamma_1$ and $\gamma_2$, then all edge energies $E_{ij}$ in problem~\eqref{eq:genMove} are submodular.
\end{proposition}
\begin{proof}
It is sufficient to show this for an arbitrary $E_{ij}$ and any possible assignment to $x_i$ and $x_j$.  If $x_i\neq\alpha$ and $x_j\neq\alpha$, then we require $E_{ij}(\alpha,\alpha) + E_{ij}(x_i,x_j) \leq E_{ij}(x_i,\alpha) + E_{ij}(\alpha,x_j)$ 
as with $\alpha$-expansion moves, which is~\eqref{eq:triangle} with $\gamma_1=x_i$ and $\gamma_2=x_j$.
If $x_i=\alpha$ and $x_j=\alpha$, then we require $E_{ij}(\alpha,\alpha) + E_{ij}(\beta,\beta) \leq E_{ij}(\beta,\alpha) + E_{ij}(\alpha,\beta)$ as with $\alpha\beta$-swap moves, which is~\eqref{eq:triangle} with $\gamma_1=\beta$ and $\gamma_2=\beta$.  If $x_i=\alpha$ and $x_j\neq\alpha$, then we require $E_{ij}(\alpha,\alpha) + E_{ij}(\beta,x_j) \leq E_{ij}(\beta,\alpha) + E_{ij}(\alpha,x_j)$, which is~\eqref{eq:triangle} with $\gamma_1=\beta$ and $\gamma_2=x_j$.  The remaining possibility 
is similar.
\end{proof}
This implies that an optimal $\alpha$-expansion $\beta$-shrink move can be computed by solving a minimum-cut problem, using (for example) the construction of~\citet{kolmogorov2002energy}.
Thus, an optimal $\alpha$-expansion $\beta$-shrink move can be computed in polynomial time under the same condition required to use $\alpha$-expansions.  Further, computing an optimal $\alpha$-expansion move and computing an optimal $\alpha$-expansion $\beta$-shrink move have the same worst-case time-complexity.
This would indicate that the new moves could be used in place of $\alpha$-expansions in any of the many applications where these moves are currently used.

\subsection{Problems with Many States}

In the computer vision problems that originally motivated the moves based on minimum-cuts, the number of states $N$ may be non-trivial, since the states may represent a discretization of a continuous value.  For example,~\citet{szeliski2007comparative} evaluate the performance of these moves on problems with $256$ states, each state representing an intensity level in an image.
Thus, in addition to their better empirical performance, $\alpha$-expansions may be preferred over $\alpha\beta$-swaps simply because the number of possible values of $\alpha$ is $N$, while the number of (non-exchangeable) combinations of $\alpha$ and $\beta$ is $N(N-1)/2$.  

This quadratic scaling in terms of $N$ would also seem to be a problem for $\alpha$-expansion $\beta$-shrink moves.  However, if we consider \emph{any} strategy for choosing a value of $\beta$ given $\alpha$, then we still obtain an algorithm that dominates $\alpha$-expansions (assuming we do not always choose $\alpha=\beta$, corresponding to the special case of $\alpha$-expansions).  
Thus, the new moves can be modified to have the same scaling with $N$ as $\alpha$-expansions.  That is, if $N$ is very large we can consider variants where $\beta$ is function of $\alpha$ so that we only consider $N$ moves instead of $N^2$, and these variants would still dominate $\alpha$-expansions.
We empirically evaluate three possible strategies for selecting $\beta$ given $\alpha$ in \S\ref{sec:experiments}.
 

\subsection{Truncation for Non-Submodular Potentials}
\label{sec:truncate}

In many problems condition~\eqref{eq:triangle} is not satisfied.  In these cases, a widely-used approach is to modify the potentials to be submodular, in such a way that an optimal move with the modified energy is guaranteed to not increase the original energy~\citep{rother2005digital}.  In the case of $\alpha$-expansions, one way to construct such a modified energy is by replacing each $E_{ij}(x_i,x_j)$ with
\begin{align*}
\bar{E}_{ij}(x_i,x_j) = \min\{&E_{ij}(x_i,x_j),\\
&E_{ij}(\alpha,x_j)+E_{ij}(x_i,\alpha)-E_{ij}(\alpha,\alpha)\}.
\end{align*}
Condition~\eqref{eq:triangle} holds with this modified energy by construction.  Further, the optimal $\alpha$-expansion with this modified energy does not increase the original energy, since the modified energy simply decreases the energy of the current assignment $(x_i,x_j)$.  As discussed in~\citep{rother2005digital}, we can alternately increase $E_{ij}(x_i,\alpha)$ or $E_{ij}(\alpha,x_j)$ to make condition~\eqref{eq:triangle} satisfied while maintaining the descent property of the moves.

We can define a modified energy function with similar properties in the case of $\alpha$-expansion $\beta$-shrink moves, though maintaining the descent property requires a slightly more complicated construction.  Although there are many possible constructions, we describe one here.
If $x_i\neq \alpha$ and $x_j \neq \alpha$, then as before we take:
\begin{align*}
\bar{E}_{ij}(x_i,x_j) = \min\{&E_{ij}(x_i,x_j),\\
&E_{ij}(\alpha,x_j)+E_{ij}(x_i,\alpha)-E_{ij}(\alpha,\alpha)\}.
\end{align*}
If $x_i=\alpha$ and $x_j=\alpha$, then we take:
\begin{align*}
\bar{E}_{ij}(\alpha,\alpha)  = \min\{& E_{ij}(\alpha,\alpha),\\
& E_{ij}(\alpha,\beta)+E_{ij}(\beta,\alpha)-E_{ij}(\beta,\beta)\}.
\end{align*}
If $x_i\neq \alpha$ and $x_j = \alpha$, then we take:
\begin{align*}
\bar{E}_{ij}(\alpha,\beta) = \max\{&E_{ij}(\alpha,\beta),\\
&E_{ij}(\alpha,\alpha)+E_{ij}(x_i,\beta)-E_{ij}(x_i,\alpha)\}.
\end{align*}
Finally, if $x_i=\alpha$ with $x_j \neq \alpha$ we take:
\begin{align*}
\bar{E}_{ij}(\beta,\alpha) = \max\{&E_{ij}(\beta,\alpha),\\
&E_{ij}(\alpha,\alpha)+E_{ij}(\beta,x_j)-E_{ij}(\alpha,x_j)\}.
\end{align*}
The other terms in the energy function are unchanged.
If~\eqref{eq:triangle} is already satisfied, then the modified energy under this construction is identical to the original energy.  This construction maintains the appealing property that any move that does not increase the modified energy will not increase the original energy.

\section{Experiments}
\label{sec:experiments}

To empirically evaluate the performance of the new moves, we performed several experiments on the non-binary data sets examined by~\citet{szeliski2007comparative}.  These data sets are summarized in Table 1, and we extracted the terms in the energy functions from  the code available online.\footnote{\url{http://vision.middlebury.edu/MRF/}}

Our experiments compared the following methods:
\begin{itemize}
\item $\alpha\beta$-Swap: performing $\alpha\beta$-swap moves in the order $\beta=1,2,\dots,N$ in an outer loop and $\alpha=\beta+1,\beta+2,\dots,N$ in an inner loop.
\item $\alpha$-Expansion: performing $\alpha$-expansion moves in the order $\alpha=1,2,\dots,N$.
\item Random $\beta$: performing $\alpha$-expansion $\beta$-shrink moves in the order $\alpha=1,2,\dots,N$ with $\beta$ selected randomly among $\{1,2,\dots,N\}$.
\item $\beta = \alpha-1$: performing $\alpha$-expansion $\beta$-shrink moves in the order $\alpha=1,2,\dots,N$ with $\beta$ set to $\max\{1,\alpha-1\}$.
\item $\beta = \alpha+1$: performing $\alpha$-expansion $\beta$-shrink moves in the order $\alpha=1,2,\dots,N$ with $\beta$ set to $\min\{N,\alpha+1\}$.
\item All $\beta$: performing $\alpha$-expansion $\beta$-shrink moves in the order $\beta=1,2,\dots,N$ in an outer loop and $\alpha=1,2,\dots,N$ in an inner loop.
\end{itemize}
Note that the iterations of the first and last method are \emph{much} more expensive for large $N$ because they consider $\mathcal{O}(N^2)$ combinations of $\alpha$ and $\beta$, while the remaining methods only consider $N$ moves.

\begin{table}
\centering
\caption{Data sets from~\citet{szeliski2007comparative} used in the experiments.}
\label{tab}
\begin{tabular}{l|l|c|c|c}
Name & Task & Nodes & Edges & States\\
\hline
Family & Montage   & 425632 & 849946 & 5  \\
Pano & Montage   & 514080 & 1026609 & 7  \\
Tsukuba & Stereo   & 110592 & 220512 & 16 \\
Venus & Stereo   & 166222 & 331627 & 20  \\
Teddy & Stereo  & 168750 & 336675 & 60  \\
Penguin & Restoration   & 21838 & 43375 & 256  \\
House & Restoration   & 65536 & 130560 & 256  
\end{tabular}
\end{table}

In our first experiment, we initialized all variables to the first state and ran each method until the energy did not change between iterations.  We used the truncation described in~\S\ref{sec:truncate} for problems that did not satisfy~\eqref{eq:triangle} for all triplets of states.  In Table~2, we show the energy of the local minima obtained divided by the energy of the local minimum with respect to $\alpha$-expansion moves.  In this table, a value of $1$ indicates that the energy was identical to the energy obtained by $\alpha$-expansion moves, and we use $1.0000$ if the energy is close but not identical.  In this experiment, the new moves with all $\beta$ obtained the lowest energy on $6$ of the $7$ data sets, and strictly so in $4$ of these cases (the exceptions were \emph{Teddy} where using a random $\beta$ lead to a lower score, and the montage data sets where other methods reached the same energy).  Among the remaining methods, the more computationally efficient strategy of simply setting $\beta=\alpha+1$ obtained the lowest energy on $5$ of the $7$ data sets.

\begin{table*}
\centering
\caption{Energy of local minima with respect to iterative descent methods beginning from all variables set to state $1$, divided by the energy of the local minima with respect to $\alpha$-expansion moves.}
\begin{tabular}{l|c|c|c|c|c|c|}
Name & $\alpha\beta$-Swap & $\alpha$-Expansion & Random $\beta$ & $\beta = \alpha-1$ & $\beta = \alpha+1$ &  All $\beta$\\
\hline
Family & 1.0203 & 1 & 0.9998 & 1 & 0.9998 & 0.9998 \\
Pano & 1.3182 & 1 & 1.0006 & 1 & 1 & 1 \\
Tsukuba & 1.0315 & 1 & 1.0012 & 1 & 1.0000 & 1.0000 \\
Venus & 1.8561 & 1 & 1.0015 & 0.9992 & 0.9979 & 0.9968 \\
Teddy & 1.0037 & 1 & 0.9998 & 1 & 1.0007 & 0.9999 \\
Penguin & 1.1283 & 1 & 1.0037 & 0.9936 & 0.9793 & 0.9758 \\
House & 0.7065 & 1 & 0.7841 & 0.9973 & 0.7038 & 0.7032 \\
\end{tabular}
\end{table*}

\begin{figure*}
\label{fig:house}
\includegraphics[width=.33\textwidth]{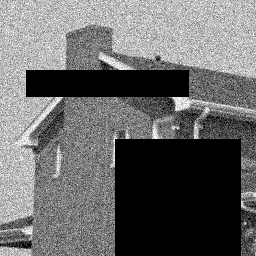}
\includegraphics[width=.33\textwidth]{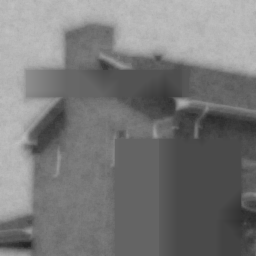}
\includegraphics[width=.33\textwidth]{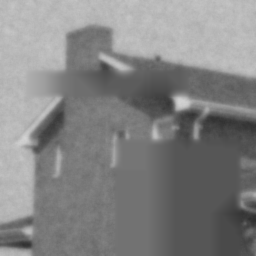}
\caption{From left to right: Initial degraded image with missing areas, local minimum of energy function with respect to $\alpha$-expansions, improved local minimum found by $\alpha$-expansion $\beta$-shrink moves with $\beta=\alpha+1$.}
\end{figure*}

In our second experiment we used the local minimum with respect to $\alpha$-expansion moves as the initialization, and tested whether the methods that only consider $N$ moves could escape from this local minimum.  Table~3 shows the energies of the local optima obtained with this initialization.  Here, we see that even with these simple choices of $\beta$ that the new moves are able to escape the local minimum with respect to $\alpha$-expansions for $5$ out of the $7$ data sets.  In this experiment choosing $\beta=\alpha+1$ gave the lowest energy on all data sets, and gave a strictly lower energy on $4$ of the $7$.  Selecting $\beta=\alpha+1$ is somewhat intuitive, the moves can be viewed as `prematurely' expanding the next value of $\alpha$ into the region occupied by the current value of $\alpha$.  
Although an improved configuration was found on at least one data set for each of the three tasks, the largest improvements were seen in the image restoration/inpainting tasks.  This is likely due to the $256$ possible labels, which is much larger than the other tasks.  In Figure~\ref{fig:house}, we show the restoration of the \emph{house} data where the largest improvement was observed.  In this figure we see a noticeable difference between the local minima with respect to the different moves.  This visual difference is also reflected quantitatively; the new moves lead to a reconstruction error ($\ell_1$-norm distance to the original image) that is approximately two-thirds that of the local minimum with respect to $\alpha$-expansions (reduced from 1303537 to 911207).

\begin{table}
\centering
\caption{Relative energy of local minima for different choices of $\beta$ when initializing with a local minima with respect to $\alpha$-expansions.}
\begin{tabular}{l|c|c|c|}
Name & Random $\beta$ & $\beta = \alpha-1$ & $\beta = \alpha+1$\\
\hline
Family & 0.9998 & 1 & 0.9998  \\
Pano &  1 & 1 & 1  \\
Tsukuba &  1 & 1 & 1  \\
Venus & 1.0000 & 0.9992 & 0.9979 \\
Teddy &  1 & 1 & 0.9999  \\
Penguin &  0.9998 & 0.9902 & 0.9775\\
House &   0.8050 & 0.9971 & 0.7038  \\
\end{tabular}
\end{table}

\section{Discussion}
\label{sec:discussion}

There have been a variety of other generalizations of $\alpha\beta$-swaps and $\alpha$-expansions proposed in the literature, and an exhaustive list of this literature would be outside our scope.  However, among the generalizations we are aware of these other moves either (i) require additional assumptions, or (ii) can not be solved in polynomial-time.  For example, range-swap and range-expansion moves generalize $\alpha\beta$-swaps and $\alpha$-expansions (respectively), but these can not be computed in polynomial-time without further assumptions~\citep{veksler2007graph,gould2009alphabet,kumar2011improved}.  \citet{kolmogorov2007minimizing} consider implementing minimum-cut methods wihout restrictions on the energy functions, but in general their algorithm can not compute the optimal move.  
\citet{lempitsky2010fusion} consider a general class of moves called fusion moves.  These moves consider two full configurations of the variables $x$ and $z$, and the set of moves are of the form
\[
y_i \leftarrow
\begin{cases}
\textrm{$x_i$ or $z_i$.}\\
\end{cases}
\]
That is, we can replace an arbitrary number of elements of $x$ by the corresponding elements from some alternative configuration $z$.  The $\alpha$-expansion $\beta$-shrink moves proposed here are a special case of this type of move, where $z_i$ is set to $\beta$ if $x_i = \alpha$ and $z_i$ is set to $\alpha$ otherwise.  However, while Proposition 4 shows that this special case is solvable in polynomial-time under condition~\eqref{eq:triangle},~\citet{lempitsky2010fusion} do not give restrictions that would allow an optimal general fusion move to be solved in polynomial time. Indeed, because of the generality of fusion moves, any such restriction would need to be much more restrictive than~\eqref{eq:triangle}.

Although in this paper we have focused on the case of minimizing unary and pairwise energies, it is possible to apply $\alpha\beta$-swaps and $\alpha$-expansions to certain classes of higher-order energies~\citep{kohli2007p3}.  We expect that the new moves discussed in this paper can also be extended to these scenarios, and indeed we believe that the new moves could be used in place of $\alpha$-expansions in any of the many applications where these moves are currently used.  
Finally, we note that the implementation of the new moves will be made available online.

\section*{Acknowledgements}

We would like to thank Olivier Duchenne and Daniel Tarlow for valuable discussions, and the anonymous reviewers for helpful comments that improved the paper.  Mark Schmidt is supported by the SIERRA grant from the European Research Council (SIERRA-ERC-239993).  Karteek Alahari is supported by the Quaero Programme, funded by the OSEO.
 
 {\small
\bibliographystyle{abbrvnat}
\bibliography{egbib}
}

\end{document}